\def\year{2021}\relax
\documentclass[letterpaper]{article} 
\usepackage{aaai21}  
\usepackage{times}  
\usepackage{helvet} 
\usepackage{courier}  
\usepackage[hyphens]{url}  
\usepackage{graphicx} 
\usepackage{natbib}  
\usepackage{caption} 
\def\nocopyright  

\usepackage[export]{adjustbox}
\usepackage{cite}
\usepackage{amsmath}
\usepackage{amssymb}
\usepackage{amsfonts}
\usepackage[caption=false]{subfig} 
\usepackage{algorithm}
\usepackage{algorithmic}
\usepackage[binary-units=true,separate-uncertainty=true,multi-part-units=single,per-mode=symbol]{siunitx}

\newcommand{\vx}{\mathbf{x}}
\newcommand{\vv}{\mathbf{v}}

\usepackage{tabularx}
\usepackage{booktabs}
\newcolumntype{C}{>{\centering\arraybackslash}X} 
\usepackage{color, colortbl}
\definecolor{Gray}{gray}{0.9}

\usepackage{paralist}

\DeclareMathAlphabet\mathbfcal{OMS}{cmsy}{b}{n}  

\title{%
    Predicted Composite Signed-Distance Fields for Real-Time Motion Planning \\%
    in Dynamic Environments
}
\author{
    Mark Nicholas Finean,\textsuperscript{\rm 1}
    Wolfgang Merkt,\textsuperscript{\rm 1}
    and Ioannis Havoutis\textsuperscript{\rm 1}\\%
    }
\affiliations{
    \textsuperscript{\rm 1} Oxford Robotics Institute, University of Oxford \\
    \{ mfinean, wolfgang, ioannis \}@robots.ox.ac.uk
}

\begin{document}

\maketitle
\thispagestyle{empty}
\pagestyle{empty}

\begin{abstract}
We present a novel framework for motion planning in dynamic environments that accounts for the predicted trajectories of moving objects. We explore the use of \textit{composite signed-distance fields} in motion planning and detail how they can be used to generate signed-distance fields (SDFs) in real-time to incorporate predicted obstacle motions; to achieve this, we introduce the concept of \textit{predicted signed-distance fields}. We benchmark our approach of using composite SDFs against performing exact SDF calculations on the workspace occupancy grid. Our proposed technique generates predictions substantially faster and typically exhibits an 81--97\si{\percent} reduction in time for subsequent predictions. We integrate our framework with GPMP2 to demonstrate a full implementation of our approach in real-time, enabling a 7-DoF Panda manipulator to smoothly avoid a moving obstacle in simulation and hardware. 
\end{abstract}

\section{Introduction}
To integrate autonomous systems into our daily lives, we need to ensure that they operate safely in the dynamic world around them. We must, therefore, develop robots that can adapt their movements as necessary to avoid moving obstacles. This holds whether we consider robots in a household environment avoiding humans and pets, autonomous vehicles on a road avoiding other cars and pedestrians, or robots operating in a dynamic industrial setting.  

\begin{figure}[ht]
    \centering
    \includegraphics[width=0.6\columnwidth]{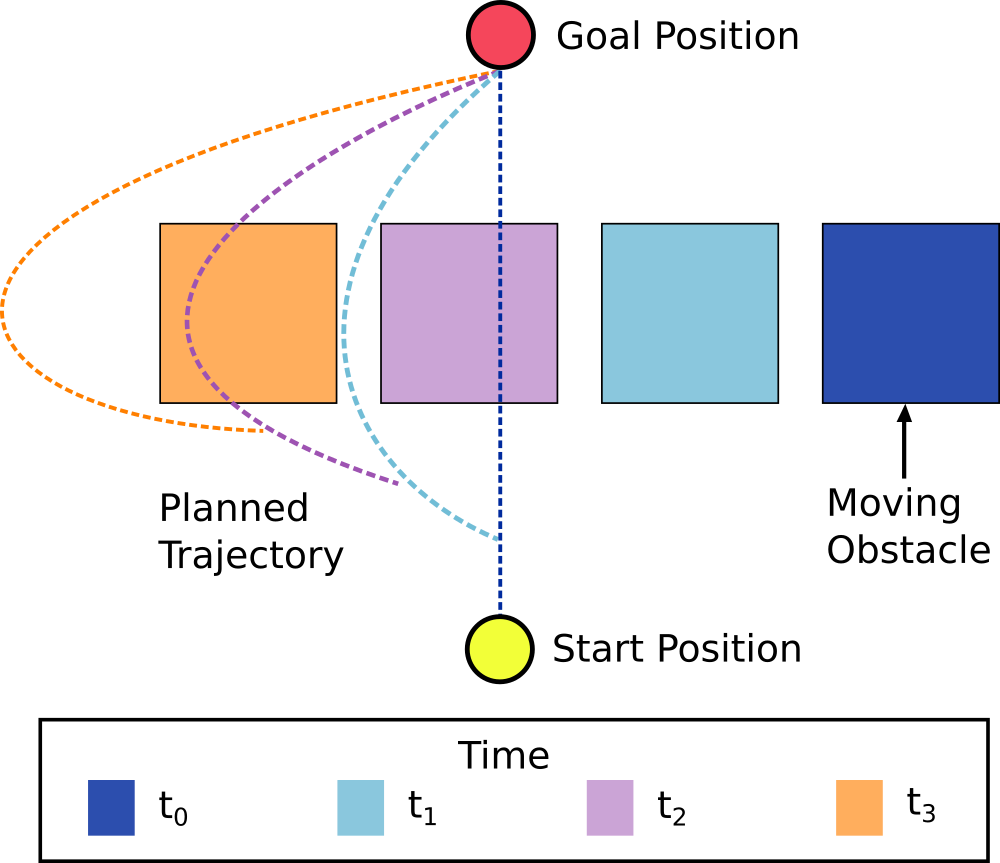}
    \caption{Toy example showing the inherent limitation of the execute-and-update method common in trajectory optimisation: The task is to plan a collision-free trajectory to reach the goal state in the presence of moving obstacles. During execution, we continue to observe the latest position of the moving obstacle and update the planned trajectory. As an obstacle approaches the intersection with the current planned trajectory, this repeated optimisation can result in the trajectory plan chasing in front of the obstacle.}
    \label{fig:execute_update_flaw}
\end{figure}

There has been significant research into motion planning; however, the current ability of robots to react in dynamic environments, with moving obstacles, is still lacking. Many approaches rely on fast re-optimisation to account for changes in the environment, yet do not account for the predicted motions of the obstacles. By neglecting predicted motions, robots encounter failure cases, as depicted in Figure~\ref{fig:execute_update_flaw}, where depending on the speed of the obstacle, the robot repeatedly plans to move in front of the moving obstacle. This scenario occurs because the optimisation can become trapped in a local minimum from which it cannot escape since it does not have full information of the obstacle trajectory during the early iterations. The resultant trajectory is both sub-optimal and potentially dangerous. In this paper, we show that the solution to this is to incorporate predicted obstacle motions into the planning problem.

\begin{figure}[t]
    \centering
    \subfloat[][\label{fig:static_traj}]{%
      \includegraphics[height=3cm,trim={8cm 0cm 12cm 0cm},clip]{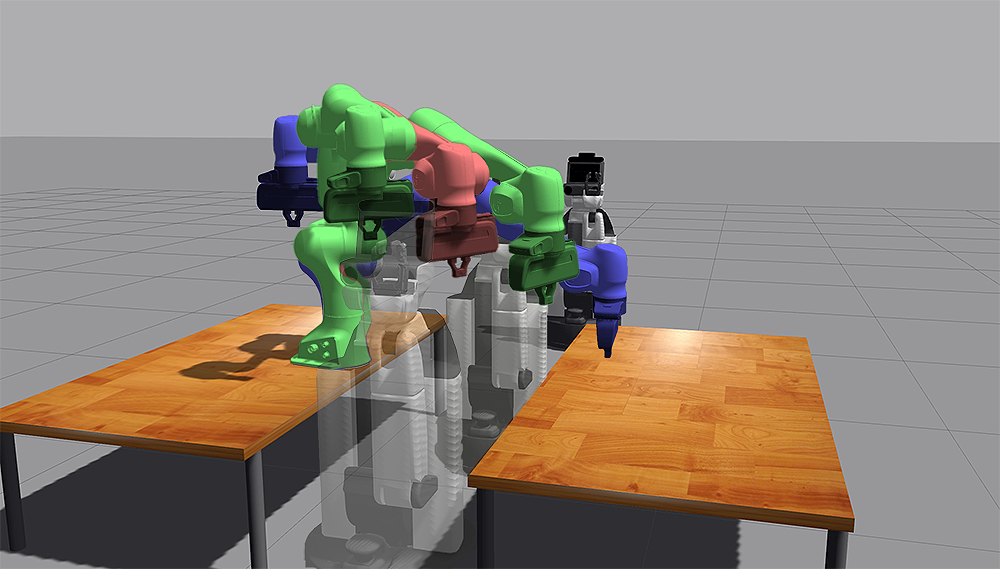}}        
      \hfill
    \subfloat[][\label{fig:one_step_traj}]{%
      \includegraphics[height=3cm,trim={0cm 0cm 12cm 0cm},clip]{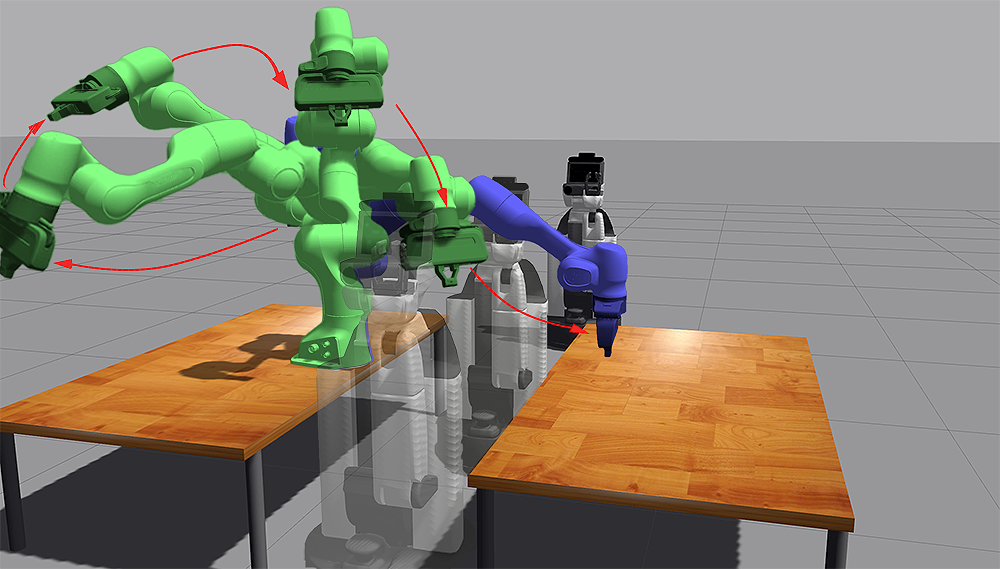}}
    \hfill
      \subfloat[][\label{fig:prediction_traj}]{%
      \includegraphics[height=3cm,trim={8cm 0cm 12cm 0cm},clip]{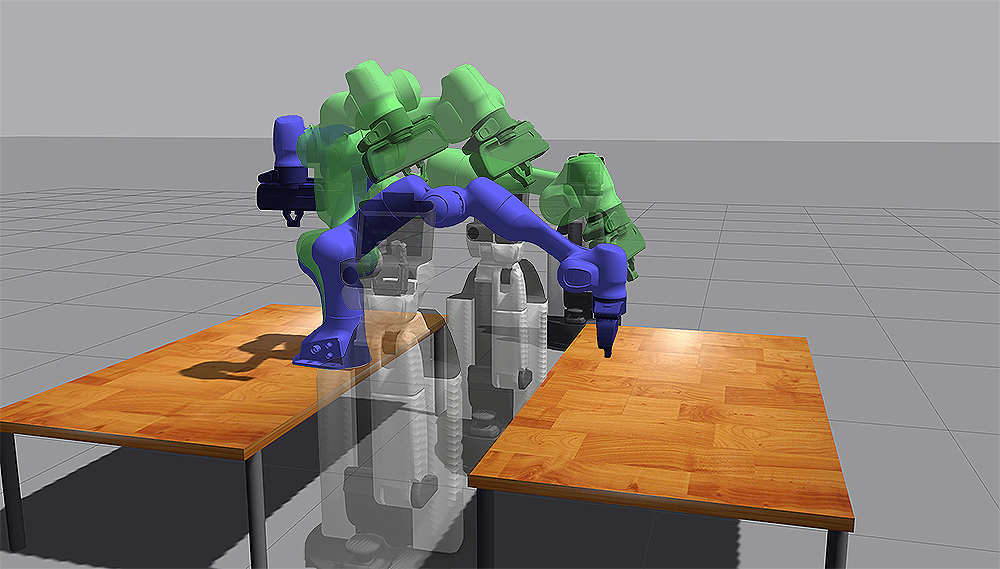}}
     \caption{Real-time collision avoidance in a dynamic environment using our proposed framework. The 7-DoF Panda arm is performing a pick-up task, positioning the end-effector over a gap onto another table. During execution, a Toyota HSR travels along the gap, acting as a moving obstacle. Poses are shown in blue (start and goal configurations), green (collision-free), or red (collision). \ref{fig:static_traj}) Without environment updates during execution. \ref{fig:one_step_traj}) With environment updates and trajectory re-optimisation during execution. \ref{fig:prediction_traj}) Our method - including obstacle trajectory prediction. We incorporate ESDF predictions which are generated at \SI{400}{\hertz}. The update loop runs at \SI{3.7}{\hertz} at the start and \SI{10}{\hertz} for the remainder of the trajectory.} 
    \label{fig:intro}
\end{figure}

Motion planners that would otherwise use signed-distance fields are forced to adopt alternative methods, such as binary collision costs \cite{itomp}, to incorporate predictions. This is because the computation time for calculating SDFs prohibits real-time usage---i.e., the ability to generate predictions at a much higher frequency than environment observations are made, typically \SI{30}{\hertz}. Our results motivate using predictions and show that we can generate real-time predictions for signed-distance fields by using \textit{composite signed-distance fields}. 

We integrate our method of composite SDF predictions into our novel framework which uses the GPMP2 motion planner \cite{gpmp2} to solve for collision-free trajectories in environments with moving obstacles. 

The key contributions of this paper are:
\begin{compactenum}
    \item Introduction of \textit{predicted signed-distance fields} for motion planning in the presence of moving obstacles.
    \item Extension of GPMP2 to plan using predicted obstacle trajectories.
    \item A novel framework for motion planning in dynamic environments using composite signed-distance fields.
\end{compactenum}

\section{Related Work}
Motion planning in the presence of moving obstacles is an open problem, while solutions generally fall into two categories. The first is to assume full prior knowledge of the moving obstacle trajectories in the scene \cite{Menon2014, Hsu2002, Merkt2019}. The second approach is `continuous re-planning' in which the motion planner either re-optimises and adapts the current planned trajectory, or considers multiple trajectory modes at any one time, such as in ITOMP \cite{itomp}, and smoothly switches between them as new information is provided \cite{Kolur2019, Zucker2013}. However, there have been few works that incorporate predicted obstacle motions into the motion planning of articulated systems.

Nam \textit{et al.} model obstacle motions as a random walk process and assign an artificial potential based on the probability density function of their predicted positions \cite{YunSeokNam1996}. Park \textit{et al.} proposed IAMP which integrates ITOMP with human motion prediction based on supervised learning \cite{Park2017}. Li and Shah implement a probabilistic roadmap approach in which the roadmap is constructed using an obstacle motion prediction, essentially assuming a prior knowledge of the trajectory \cite{Li2019}.

To further explore this problem, we first consider the environment representations used in motion planners.

Binary occupancy information enables search-based methods such as A* and D*-Lite, as well as sampling-based methods like Probabilistic Road Maps (PRMs) and Rapidly-exploring Random Trees (RRTs), to perform collision avoidance \cite{Hart1968,Koenig2002,Kavraki1996, Kavraki1998,LaValle1998,Kuffner2000}. In contrast, continuous optimisation-based approaches additionally require gradients so commonly use a Euclidean distance transform (EDT), or Euclidean signed-distance field (ESDF), to represent the environment.

When planning in 3D space, the environment is most often split into an array of discrete volume elements. We define a `voxel' to represent a single discrete volume in space. An ESDF is a 3D voxelgrid in which each cell contains the signed-distance to the closest obstacle surface. The sign denotes whether the cell is within or outside the surface boundary. In addition to providing distance information which can be used to assign collisions costs, tri-linear interpolation enables us to obtain gradients.

Motion planners such as CHOMP, TrajOpt, and GPMP2 use ESDFs to represent the environment and demonstrate them to be very effective in enabling fast planning in static environments \cite{chomp,trajopt,gpmp2}. However, the limitations of SDFs become apparent when we look to apply them in dynamic environments. 

ITOMP considered obstacle trajectory prediction for use in dynamic environments. However, in contrast to their general approach for static collision costs, which uses EDTs, they implement a simple occupancy cost for predicted motions based on ``geometric collision detection between the robot and moving obstacles" \cite{itomp}. At each time-step in the optimisation a binary cost is allocated for every moving obstacle that is in collision. The reason for their approach is that EDTs are commonly considered to have CPU compute times that are not fast enough for real-time performance, thus are pre-computed and assumed to be static. 

Various methods have been proposed to reduce the compute time for exact EDTs. Maurer \textit{et al.} show that an exact distance transform can be calculated in linear time \cite{Maurer}. Approximate methods, such as the Chamfer Distance Transform (CDT) and Fast Marching Method (FMM), have also been proposed \cite{Jones2006} to achieve fast approximations. In robotics, we often accumulate information over time; this is in contrast to areas such as medical image processing, where an ESDF will be constructed from all of the information available. This has lead to significant work being conducted into faster integration of new occupancy information into signed-distance fields. Oleynikova \textit{et al.} found that Truncated Signed-Distance Fields (TSDFs) are faster to construct than Octomaps, a widely used package that maps the environment using octrees \cite{octomap}, and proposed, VoxBlox, a method to build an ESDF from projective TSDFs incrementally \cite{voxblox}. In contrast to ESDFs, TSDFs use a `projected distance' metric and consider distances only within a truncated radius of surface boundaries. TSDFs are commonly used in vision for surface reconstruction \cite{Newcombe2011, Whelan2012}. 

Despite the speed-ups achieved by state-of-the-art methods, they are not able to calculate SDFs fast enough for use in a real-time motion planner that uses what we term \textit{`predicted signed-distance fields'} - predictions of what the ESDF of the workspace will be for future times based on estimates of obstacle geometry, velocity, and direction. We propose using \textit{composite signed-distance fields} to generate predicted signed-distance fields in real-time.

Zucker \textit{et al.} first mentioned the concept of composite SDFs, noting that distance fields are compositional under the \texttt{min} operation \cite{Zucker2013}. Thus the distance field computation can be reduced to a minimisation across a set of pre-computed distance field primitives. In the computer graphics community, composition of object-centric SDFs is used to unionise objects in `Constructive Solid Geometry' as well as to combine the SDFs of individual objects into a single SDF of the scene for ray-marching \cite{Reiner2011}.  

Zucker \textit{et al.} use composite SDFs in `CHOMP-R' to quickly update their environment representation in response to a changing environment. However, to our knowledge, they have not been used in a predictive manner. We provide further analysis of this technique and leverage it in our novel framework which extends it to the domain of integrating predicted object trajectories into motion planning. 

To demonstrate our framework, we integrate it with GPMP2 which formulates motion planning as probabilistic inference on a factor graph and uses Gaussian Processes (GPs) to represent continuous-time trajectories with a small number of states \cite{gpmp2}. Trajectories are found by using the GTSAM framework \cite{GTSAM} to exploit sparsity in the problem and perform fast probabilistic inference on factor graphs. Collision avoidance is applied via the use of `obstacle factors', which evaluate the collision-free likelihood, using a hinge-loss obstacle cost for each time-step. The hinge-loss is calculated, in both CHOMP and GPMP2, by using an approximate robot model, comprised of spheres, to query the signed-distance field \cite{chomp, gpmp2}. However, Mukadam \textit{et al.} use a single pre-computed SDF of the surrounding environment that is shared by all of the obstacle factors \cite{gpmp2} and so does not incorporate the dynamics of moving obstacles. Kolur et al. consider dynamic environments but update all of the obstacle factors to use the same updated SDF and then re-optimise to solve for an updated trajectory plan \cite{Kolur2019}.

\section{Methods}
Part \ref{sect:prediction} describes our method of predicting trajectories for dynamic obstacles.
%
Part \ref{sect:sdfs} details our novel framework for predicting environment representations, using composite signed-distance fields, and then outlines how they can be integrated with a motion planner. We describe an experiment to examine the speed-up attained by using composite signed-distance fields in contrast to exact computations.
In Part \ref{sect:dynamic_gpmp2}, we describe how we adapted the GPMP2 motion planner to react in dynamic environments as well as an experiment which demonstrates its applicability.
In Parts \ref{sect:implementation} and \ref{sect:real}, we demonstrate our fully integrated framework in simulation and on hardware, respectively. In both, a 7-DoF Panda manipulator successfully avoids a moving obstacle.

\subsection{Obstacle Trajectory Prediction} \label{sect:prediction}
To demonstrate that our approach enables the incorporation of obstacle trajectory predictions, we consider a simple prediction strategy for the occupancy of the workspace. Consider a 3D workspace, $\mathcal{W}$, which comprises of a set of $i = 1,\ldots,n$ stationary objects, $\mathcal{O}^{s}_{i}$, and $j = 1,\ldots,m$ moving (dynamic) objects, $\mathcal{O}^{d}_{j}$. At a given time, $t$, a moving object with a centroid position, $\mathbf{x}_j(t)$, will travel at velocity $\mathbf{v}_{j}(t)$.  We write the collection of discretised positions and velocities as $\mathbf{X}(t)=[\vx_1(t),\vx_2(t),\ldots,\vx_m(t)]^T$ and $\mathbf{V}(t)=[\vv_1(t), \vv_2(t), \ldots, \vv_m(t)]^T$.

In our simulated experiments, we assume noiseless sensor data and subsequently an accurate occupancy grid of the workspace, $\mathcal{G}(t)$, at each time-step. At any given time, we can identify and isolate occupied regions of the workspace. From these regions, we calculate the centroid and list of voxels associated with each object. By comparing two successive frames, at times $t$ and $t+ \delta t$, we identify which objects are moving and estimate their respective velocities from the motion of each centroid---this enables us to identify occupancy grids associated with the static scene, $\mathcal{G}_{static}$, and the moving obstacles, $\mathcal{G}_{moving}(t)$.

With centroid and velocity information, we propagate each object using a constant-velocity model to produce a predicted trajectory of the centroid. This method inherently gives accurate trajectory predictions when an object is moving in a straight line with constant-velocity. However, we emphasise that the prediction model is fully replaceable and that a more sophisticated object tracking and trajectory prediction can be implemented within our framework, e.g. a KLT tracker or Unscented Kalman Filter \cite{JianboShi1994,Raj2016}. 

\subsection{Composite Signed-Distance Field Generation} \label{sect:sdfs}
\begin{figure*}[htbp]
    \centering
    \subfloat[][\label{fig:occupancy_t0}]{%
       \includegraphics[width=0.28\columnwidth]{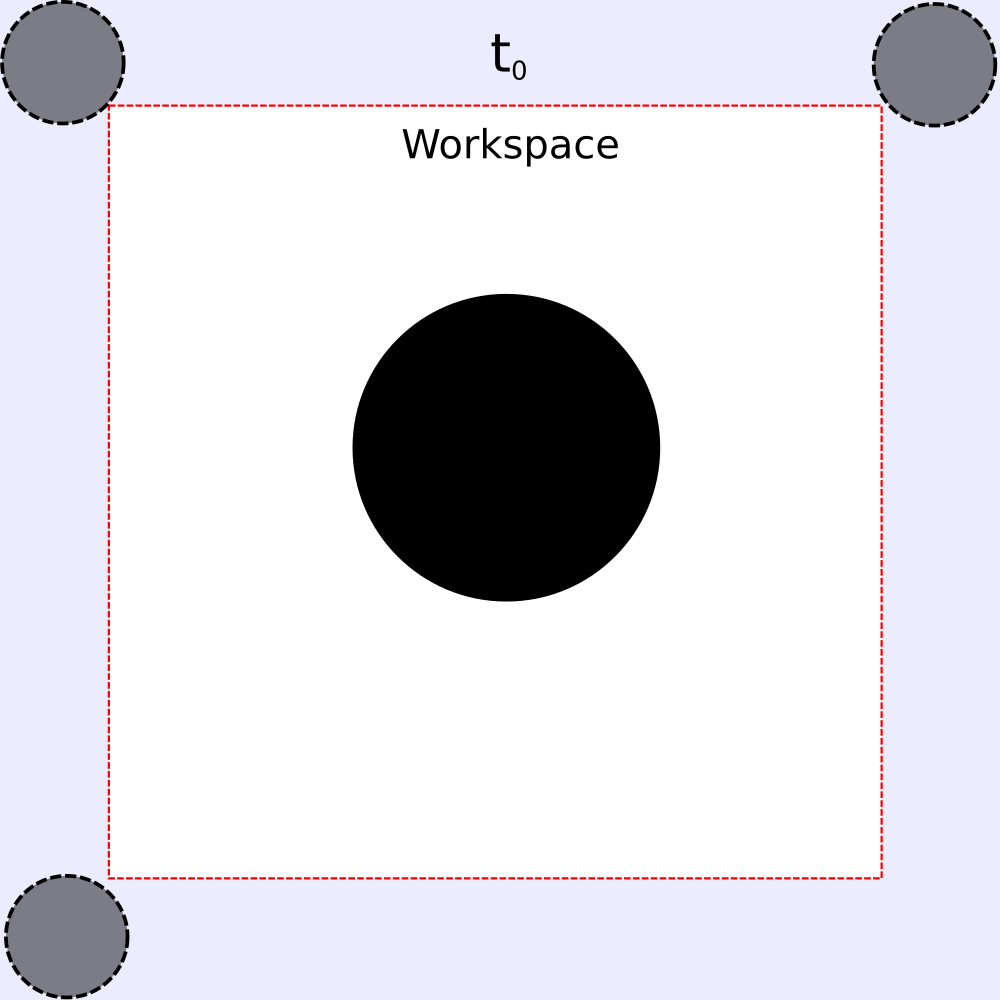}}
    \hspace{0.025\columnwidth}
    \subfloat[][\label{fig:occupancy_t1}]{%
        \includegraphics[width=0.28\columnwidth]{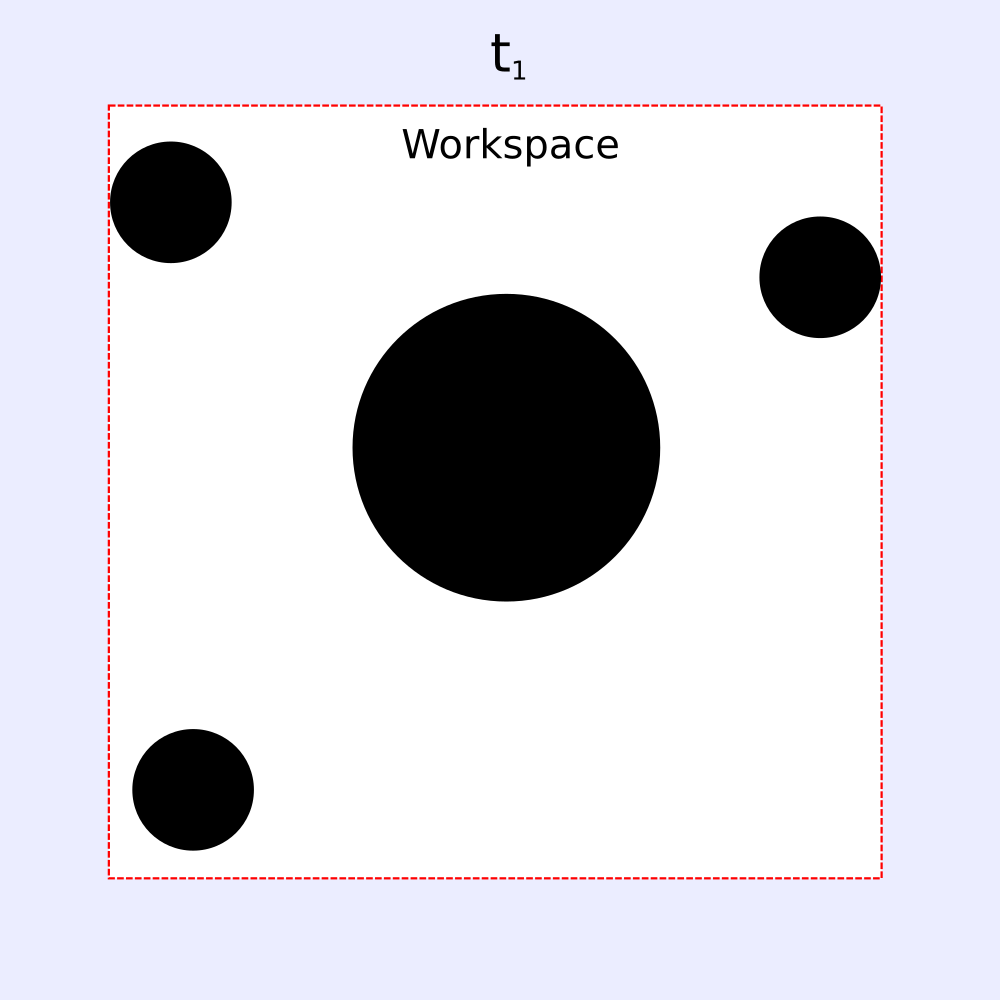}}
    \hspace{0.012\columnwidth}
    \subfloat[][\label{fig:occupancy_t2}]{%
        \includegraphics[width=0.28\columnwidth]{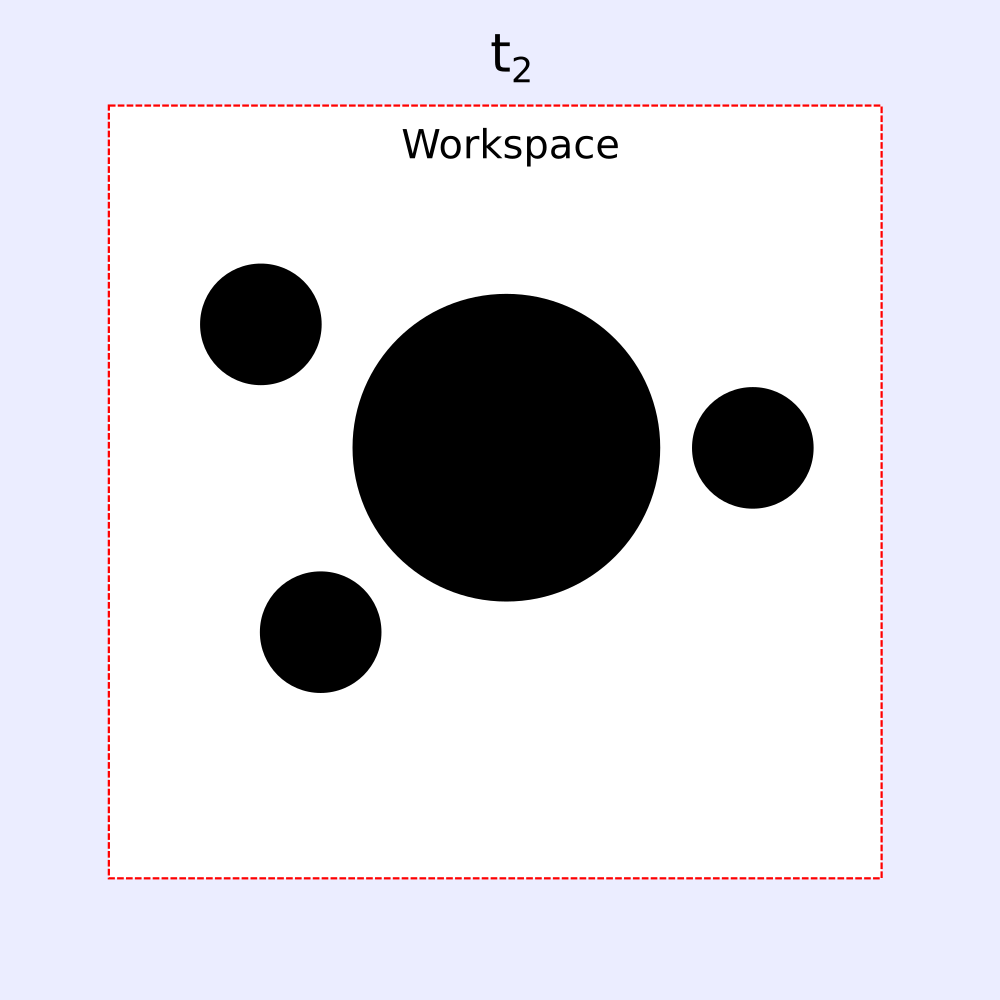}}
    \hspace{0.06\columnwidth}
    \subfloat[][\label{fig:sdf_breakdown}]{%
       \includegraphics[width=0.4\columnwidth]{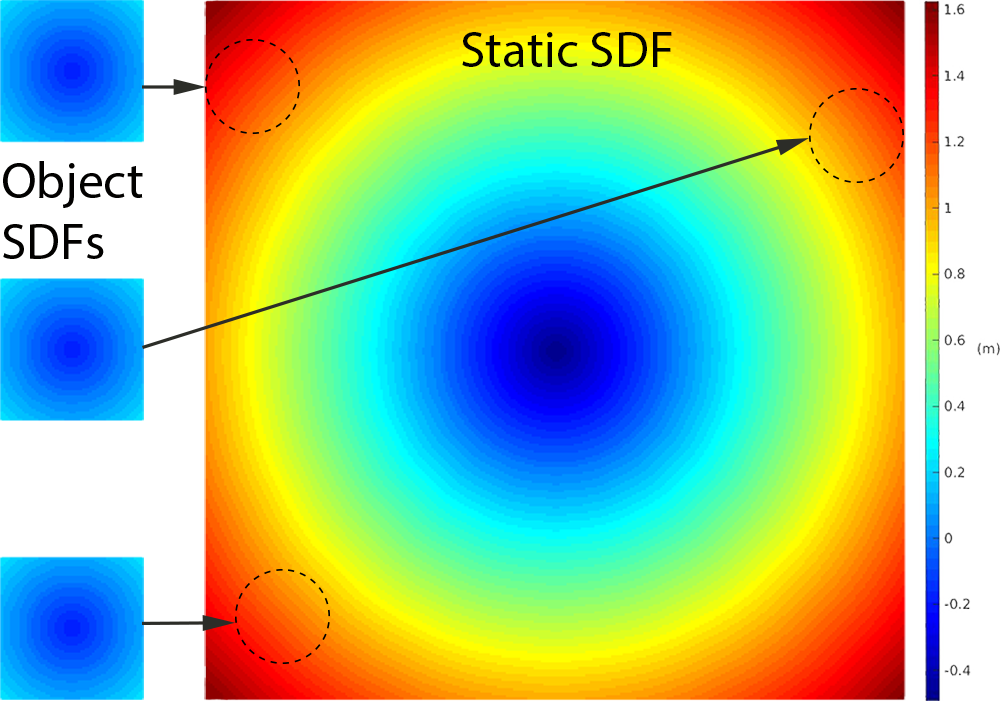}}
    \subfloat[][\label{fig:composite_sdf}]{%
        \includegraphics[width=0.28\columnwidth, ,trim={0cm 0cm 1.25cm 0cm},clip]{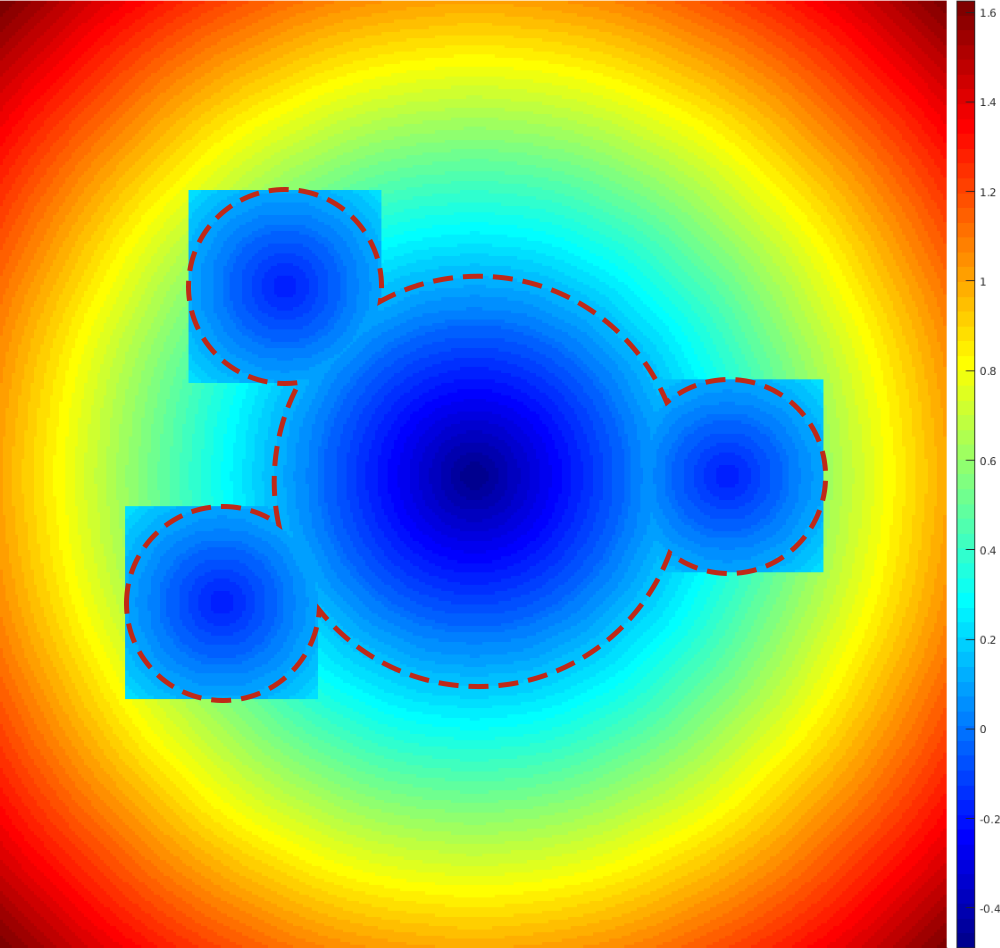}}
    \hspace{0.01\columnwidth}
    \subfloat[][\label{fig:exact_sdf}]{%
        \includegraphics[width=0.28\columnwidth,trim={0cm 0cm 1.25cm 0cm},clip]{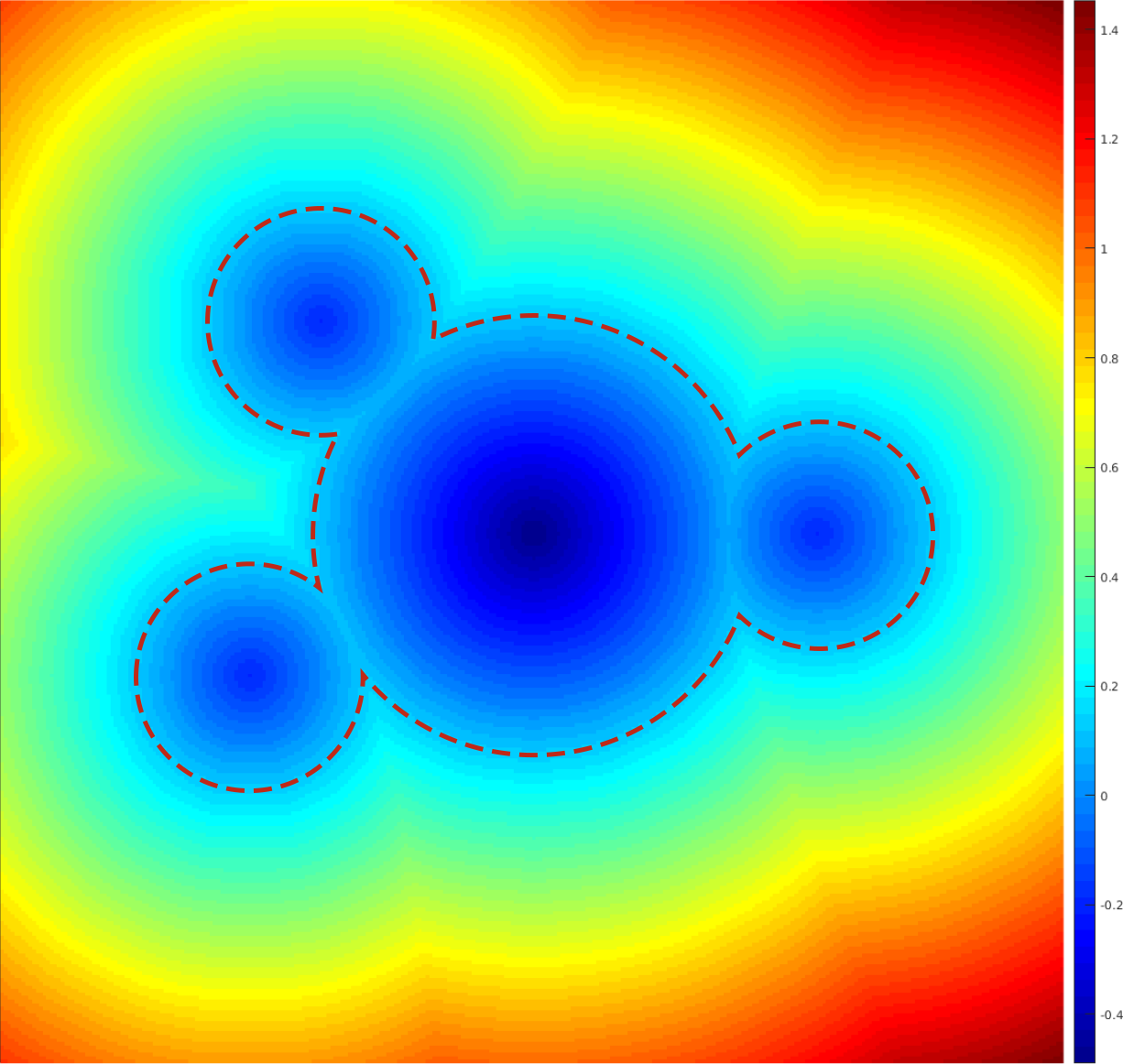}}

  \caption{\ref{fig:occupancy_t0}, \ref{fig:occupancy_t1} and \ref{fig:occupancy_t2} depict a toy example of three moving spheres entering the workspace in which a static obstacle (large central sphere) is present. White regions in the workspace represent free space, while black regions are occupied. \ref{fig:sdf_breakdown} shows an SDF of the static environment, with the tracked objects in \ref{fig:occupancy_t1} overlaid for illustration purposes. For each of the tracked objects in the scene, an SDF is calculated and associated with them. By tracking the positions of the moving objects, we infer their velocities in order to make predictions of their future positions. At these future positions, we can superpose the object SDFs onto the static SDF using a \texttt{min} operation---the result is a composite SDF. \ref{fig:composite_sdf} shows a composite SDF for $t_{2}$. For comparison, \ref{fig:exact_sdf} shows the corresponding exact SDF. Critically for motion planning, the two are identical for distances up to $\epsilon$ away from the obstacle surface boundaries, as indicated by the red dashed line.}
  \label{fig:sdfs} 
\end{figure*}
Building on the previous description of how to identify the static and moving parts of the scene, we use $\mathcal{G}_{static}$ to calculate the static SDF $\mathcal{S}_{static}$. Next, we consider an occupancy box, $\mathcal{B}_{j}$, around each of the $m$ moving objects, $\mathcal{O}^{d}_{j}$, that we have identified in $G_{moving}$. Each occupancy box is chosen to be of equal size to $\mathcal{O}^{d}_{j}$, plus an additional $\epsilon$ in all directions. For each $\mathcal{B}_{j}$, we calculate the corresponding object SDF, $\mathcal{S}_{j}$. The resultant SDF is the smallest cuboid which encloses all voxels in 3D space with signed-distance values equal to or less than a distance $\epsilon$ away from the surface of the object. We introduce $\epsilon$ as a safety margin desired to be penalised in collision avoidance. When this method is integrated with a motion planner, $\epsilon$ should equal the desired safety margin around objects to be penalised plus the size of the largest collision sphere associated with the approximate robot model.

Similar to $\mathbf{X}(t)$ and $\mathbf{V}(t)$, we denote the stacked vector of moving SDFs at a given time as $\mathbfcal{S}(t)$. In practice, we may accumulate more information about the shape of each moving object, hence $\mathbfcal{S}(t)$ is a function of time. However, for demonstration, our simulations assume noiseless sensor data and so do not acquire additional shape information in subsequent steps; we thus refer to $\mathbfcal{S}(t)$ simply as $\mathbfcal{S}$.

\begin{algorithm}
	\caption{Real-time Update Loop}
	\begin{algorithmic}[1]
		\renewcommand{\algorithmicrequire}{\textbf{Input:}}
		\renewcommand{\algorithmicensure}{\textbf{Usage:}}
		\ENSURE  Continuously updates the motion planner 
		\\ \textit{Initialisation} :
		\STATE $\mathbfcal{S} = getObjectSDFs()$
		\STATE $\mathcal{S}_{static} = getStaticSDF()$
		\\ \textit{Execution :}
		\WHILE{not shutdown}
		\STATE $\mathbfcal{S} = updateObjectSDFs()$				
		\STATE $\mathbf{X} = updateObjectPositions()$
		\STATE $\mathbf{V} = updateObjectVelocities()$
		\FOR {$t$ in $times\_to\_predict$}
	    \STATE $\mathbf{X}_{predicted} = calcFutureObjPositions(t)$
	    \STATE $\mathcal{S}_{predicted} = predictSDF(t, \mathbf{X}, \mathbf{V}, \mathbfcal{S}, \mathcal{S}_{static})$
	    \STATE $updateMotionPlanner(t, \mathcal{S}_{predicted})$	    
		\ENDFOR
	    \STATE $reoptimiseAndSendTrajectory()$
		\ENDWHILE
	\end{algorithmic}
	\label{algo:1}
\end{algorithm}
\begin{algorithm}
	\caption{$predictSDF(t, \mathbf{X}, \mathbf{V}, \mathbfcal{S}, \mathcal{S}_{static})$}
	\begin{algorithmic}[1]
		\renewcommand{\algorithmicrequire}{\textbf{Input:}}
		\renewcommand{\algorithmicensure}{\textbf{Output:}}
		\ENSURE  $\mathcal{S}_{predicted}$
		\STATE $\mathcal{S}_{predicted} = \mathcal{S}_{static}$				
		\FOR {$j = 1:m$}
		\STATE $inds = getOccupiedIndices(t, \mathbf{x}_{j}, \mathbf{v}_{j})$				
	    \STATE $\mathcal{S}_{predicted}(inds) = min(\mathcal{S}_{predicted}, \mathcal{S}_{j})$		
		\ENDFOR
		\RETURN $S_{predicted}$
	\end{algorithmic}
	\label{algo:2}
\end{algorithm}

For any future time, $t_{f}$, we can predict the occupancy of the workspace by propagating the current position $\mathbf{x}_{j}(t_{n})$ of each moving obstacle SDF $\mathcal{S}_{j}$ to its predicted position using velocity $\mathbf{v}_{j}(t_{n})$. Similar to considering an occupancy grid as the superposition of all occupied regions, we consider the workspace SDF as the superposition of all object SDFs in the scene---this forms the basis of a composite SDF. 

To calculate the resultant composite SDF for a given time, we take the minimum voxel value (i.e., signed-distance) of all overlapping scene and object SDFs. By construction, the composite SDF is guaranteed to be accurate up to the distance $\epsilon$. We illustrate this process in Figure~\ref{fig:sdfs}.

Algorithms \ref{algo:1} and \ref{algo:2} describe how we use this method in an update loop for motion planning with SDF predictions generated in real-time. While obstacle prediction in motion planning is not novel, to our knowledge this has not been performed by predicting the SDF of the environment for different time-steps to solve for a trajectory.
Our approach thus relates to time-configuration space planning, explicitly encoding the state of the environment at different points in time \cite{Merkt2019}. 

\begin{figure}
    \centering
    \subfloat[][\label{fig:moving_lab_env}]{%
       \includegraphics[width=0.46\columnwidth]{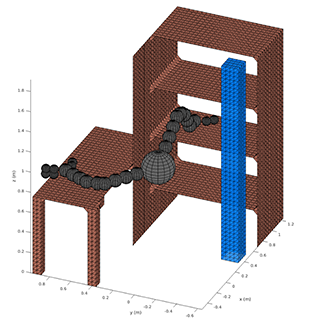}}
    \hspace{0.05\columnwidth}
    \subfloat[][\label{fig:moving_lab_env_aerial}]{%
        \includegraphics[width=0.46\columnwidth]{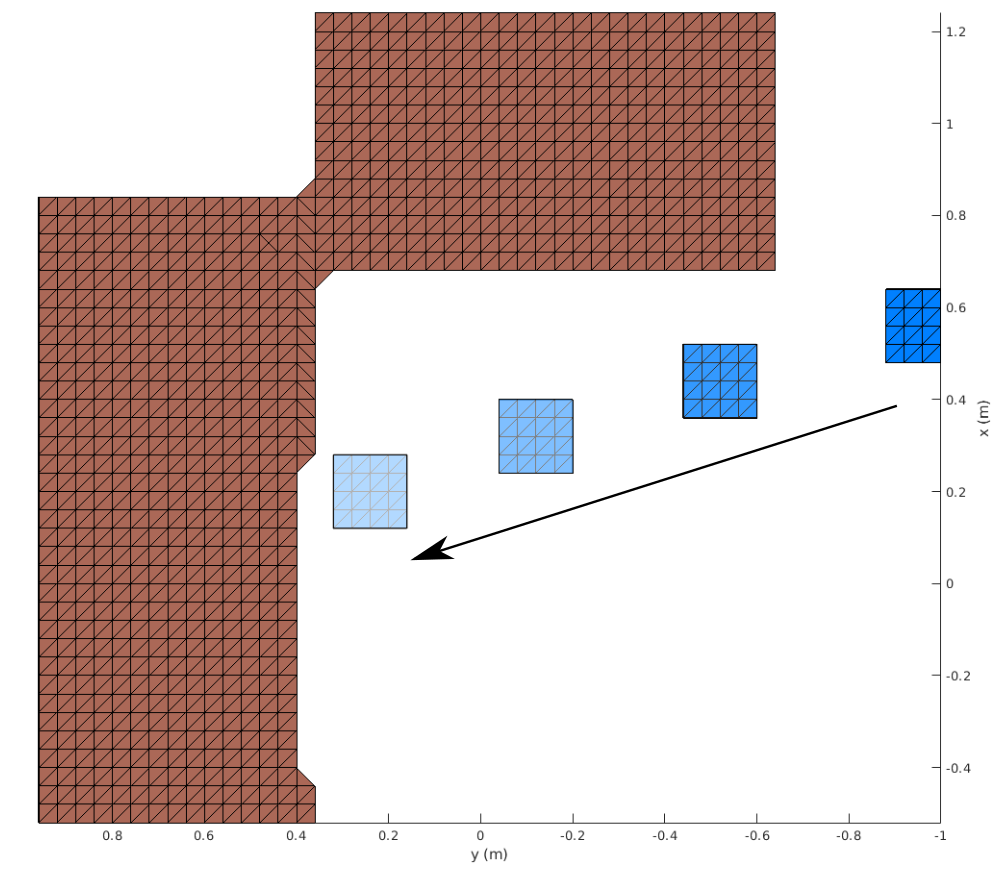}}
    \caption{`Single moving pillar' environment - Aerial and three-quarter perspective views of an example environment in which a tall pillar traverses the floor. Example start and goal configurations are depicted using a sphere approximation of the WAM robot.}
    \label{fig:envs}
\end{figure}

To evaluate the speed-up achieved using composite SDFs, we constructed a set of four dynamic datasets modelled on the `table and cabinet' setup presented in \cite{Mukadam2016}. Each of the environments contains a table and cabinet with the addition of: a) one small moving box, b) two small moving boxes, c) one moving pillar, d) two moving pillars. Figure~\ref{fig:envs} shows aerial and three-quarter perspective views of the `single moving pillar' setup. We also include a simpler environment consisting of an empty workspace with a large moving block.

For each dynamic environment, we simulate 31 different time-steps in a $96 \times 96 \times 96$ workspace with \SI{4}{\centi\metre} resolution. At each time-step, we produce composite SDF predictions for the remaining time-steps. We perform the same simulation and benchmark against the exact SDF computations as calculated in GPMP2, using MATLAB's \texttt{bwdist} function, on the predicted occupancy grids.

\subsection{Dynamic GPMP2} \label{sect:dynamic_gpmp2}
We adapt the original GPMP2 implementation to enable rapid updates of the SDF associated with each obstacle factor \textit{independently}. The process of quickly updating obstacle factors facilitates a motion planner that can re-optimise and adapt planned trajectories to a changing environment.

To examine the effect of using independent SDFs for each obstacle factor, and to assess the importance of incorporating obstacle dynamics, we considered and compared three different scenarios in dynamic environments. Each scenario considers a different degree of perception; they are as follows:
\begin{compactenum}
	\item \textit{Static:} Assumes a static workspace and all obstacle factors remain unchanged throughout the execution of the trajectory. The factor graph is thus optimised only once.
	
	\item \textit{Execute-and-update:} Obstacles factors are initialised with the initial SDF of the scene. During execution, at each time-step, all obstacle factors with a time-index greater than or equal to the current time are updated to use the current observation of the workspace. The trajectory is re-optimised and executed at each time-step in an iterative process.
	
	\item \textit{Full prior knowledge:} Each obstacle factor is provided with the known SDF of the environment at the associated time. The factor graph has maximum knowledge of the environment evolution and is thus only optimised once.
\end{compactenum}

\begin{figure}
    \centering
        \includegraphics[width=0.67\columnwidth]{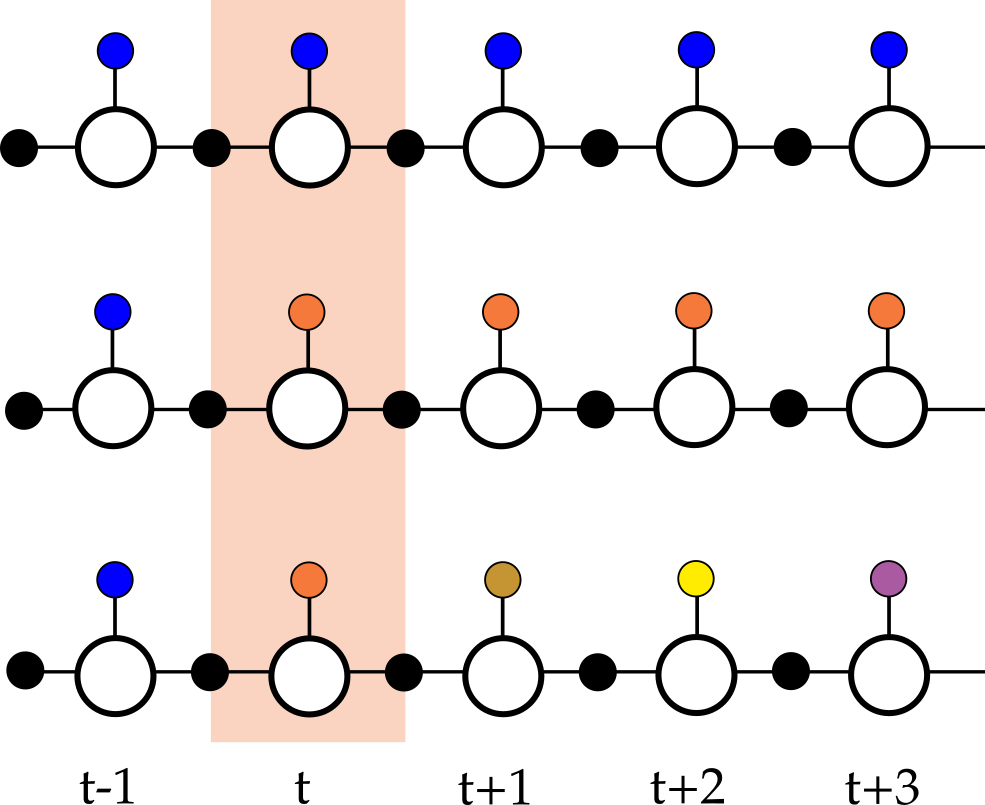}
    \caption{Factor graphs associated with three different feedback scenarios. Variable nodes are shown in white, prior factors in black, and coloured obstacle factors. \textit{Top} - the obstacle factors share the same SDF and remain unchanged during execution. \textit{Middle} - as the trajectory is executed, the current and all future obstacle factors are replaced with the latest SDF. \textit{Bottom} - as the trajectory is executed, the current obstacle factor is updated to the latest observed SDF while all future obstacle factors receive updated SDF predictions.}
    \label{fig:three_fg}
\end{figure}

The factor graphs associated with each of the perception scenarios are illustrated in Figure~\ref{fig:three_fg}. 

To test each scenario, we used a 7-DoF WAM robot arm in simulation to plan trajectories in the `single moving pillar' setup shown in Figure~\ref{fig:envs}. Plans were conducted over seven different combinations of start and goal configuration, and a range of moving obstacle speeds, \SIrange{0.1}{1.0}{\metre\per\second}. We recorded the resultant smoothness (GP) cost associated with each trajectory and the number of collisions that the robot made with the environment.

To formalise the task, we constructed a factor graph comprising of 31 time-indexed variable nodes, separated by \SI{0.1}{\second} intervals, to form a \SI{3}{\second} time horizon. Prior factors were added to specify the start and goal configurations, and Gaussian priors used to connect variable nodes, as in \cite{gpmp2}. We assigned an obstacle factor to each variable node and interpolated obstacle factors between pairs of variable nodes with a discretisation of $\tau = 5 \si{\milli\second}$ ($n_{int} = 19$) to provide dense collision checking along the trajectory. Further parameters used were $\epsilon = \SI{0.2}{\metre}$ to encourage safe clearance of obstacles, as well as $\sigma_{cost} = 0.2$, as a result of tuning, to weight the relative costs of trajectory smoothness and collision avoidance. The environment was discretised at \SI{4}{\centi\metre} resolution in a $96 \times 96 \times 96$ voxel workspace.

We performed another similar experiment on the experimental setup shown in Figure~\ref{fig:intro}, in which a Franka Emika Panda 7-DoF robotic arm was tasked with reaching across the gap between two tables to achieve a `pick-up' goal configuration. During execution, a Toyota HSR 
traverses the walkway between the tables, acting as a moving obstacle. For this setup, we used 42 different combinations of start and goal configuration. The moving obstacle speed was chosen to be in the range of \SIrange{1.0}{2.0}{\metre\per\second}, most closely resembling walking speeds of a human. We chose $\tau = 0.05 \si{\second}$ ($n_{int} = 1$) to save compute time and because the time interval between variable nodes was already small. Other parameters used were $\sigma_{cost} = 0.05$ and $\epsilon = \SI{0.3}{\metre}$ based on cost tuning. As before, the environment was discretised at \SI{4}{\centi\metre} resolution in a $96 \times 96 \times 96$ voxel workspace. 

\subsection{Closed-loop Implementation} \label{sect:implementation}
We combined our predictive framework using composite signed-distance fields with our dynamic approach, applied within GPMP2, to solve planning problems in dynamic environments. The control loop is shown in Algorithm \ref{algo:1}. In each iteration of the control loop, we observe the environment and update the workspace occupancy grid. We populate the occupancy grid by tracking the position of model states in the Gazebo simulation and inserting occupied regions into the occupancy grid at corresponding locations.

Using the methods discussed in Section \ref{sect:prediction}, we segment and track obstacles online. A constant-velocity model is applied to each of the segmented moving objects, enabling us to predict the future locations of each object and generate composite SDFs for all obstacle factors with time-index equal to or greater than the current time. To our knowledge, this is the first work to implement and update different SDFs for each time-step as in time-configuration space planning. A prior factor is also added for the current time-index to set the current configuration of the robot. The factor graph is then re-optimised and the new trajectory is merged with the current trajectory being executed on the robot.

We demonstrate our closed-loop implementation of the proposed framework on a Panda 7-DoF robotic arm in Gazebo, 
using the same task as described previously in which the Panda reaches across a gap to place the end-effector over a table, as in a pick-and-place task. We set the HSR base movement speed at \SI{1.4}{\metre\per\second} to test against a speed comparable with the average walking pace \cite{walkingspeed1, Nolan191, walkingspeed2}.

\subsection{Hardware Experiments} \label{sect:real}
We validate our dynamic GPMP2 approach in the presence of uncertainty in the execution of the trajectory, as well as the pose of the dynamic obstacle, in hardware experiments. We use a Franka Emika Panda arm in conjunction with a Realsense D435i RGB-D camera for object tracking and detection. We used \textit{apriltag\_ros}~\cite{Malyuta2019} to identify the pose of an AprilTag~\cite{Wang2016} fiducial marker affixed to a dynamic obstacle (box). For safety purposes, we used a longer time-horizon of \SI{5}{\second} and an epsilon value of \SI{0.4}{\metre} to encourage both lower joint speeds and comfortable collision avoidance. The robot was positioned on a single table and tasked with a variety of motions, including the same motion as in the previous experiment, and avoiding the box as moved by the operator. 

\section{Evaluation}
\begin{table*}[t]
\small
\caption{Signed-Distance Field Timing Analysis}
\label{table:sdf_timings}
\begin{tabularx}{\textwidth}{| c | C | C | C | C | C | C | C | C | C | C |}
\multicolumn{1}{ c }{}&\multicolumn{9}{ c }{} \\ \cline{2-10}  
\multicolumn{1}{ c |}{}&\multicolumn{9}{ c |}{Workspace Size (Voxels Per Side)}\\
\hline
                            & 64 & 96 & 128 & 160 & 192 & 224 & 256 & 288 & 320 \\ 
\hline
\rowcolor{Gray}
Composite Init. (ms)  & $13\pm1$ &  $31\pm4$ &  $65\pm3$ &  $126\pm6$ &  $211\pm3$ &  $329\pm6$ &  $475\pm7$ &  $659\pm5$ &  $936\pm23$\\
Full Computation (ms)          & $9.3\pm1.0$ & $28\pm2$ & $62\pm3$ & $121\pm5$ & $	217\pm11$ & $	338\pm9$ & $493\pm13$ & $687\pm18$ & $960\pm28$\\ 
Composite Prediction (ms)   & $0.6\pm0.6$ &  $1.9\pm0.2$ & $4.0\pm0.9$ & $7.4\pm0.5$ & $24\pm6$ & $42\pm2$ & $62\pm3$ & $86\pm3$ & $121\pm3$ \\
\hline
Repeat Prediction Speed-Up   & \textbf{15.5x} & \textbf{28.0x} & \textbf{15.5x} & \textbf{16.4x} & \textbf{9.0x} & \textbf{8.0x} & \textbf{8.0x} & \textbf{8.0x} & \textbf{7.9x} \\
\hline
\end{tabularx}
\end{table*}

We conducted experiments using an 8-core Intel Core i7-9700 CPU @ \SI{4.50}{\giga\hertz} and \SI{2133}{\mega\hertz} DDR4 RAM. 
Signed-distance field computations were performed in series, as in the original GPMP2 implementation, using the \texttt{bwdist} function in MATLAB.
GTSAM was configured to use multi-threading enabled for linearising the factor graph.

\subsection{Composite SDF Prediction}
Computation times for composite signed-distance fields and full computation from $96 \times 96 \times 96$ occupancy grids are shown in Table \ref{table:sdf_timings}. 

The composite initialisation time, highlighted in grey, comprises of the time to calculate the SDFs associated with the static workspace, $\mathcal{S}_{static}$, and each of the moving obstacles in the scene, $\mathbfcal{S}$. This initial time penalty must be paid at the startup of the experiment or any subsequent time you wish to update the object and static environment SDFs. However, our results show that beyond this initialisation penalty, it is significantly faster to generate predicted composite SDFs than to perform full SDF calculations; we see a \SIrange{7.9}{28.0}{x} speed-up for subsequent predictions depending on the discretisation of the workspace. This has a significant impact on enabling SDF predictions to be generated in real-time---conducting our experiment for a $96 \times 96 \times 96$ discretisation of the workspace, as in our closed-loop implementation, gives a mean SDF generation rate of \SI{540}{\hertz}, compared with \SI{36}{\hertz} for the benchmark.

\subsection{Dynamic GPMP2}
\begin{figure*}[htbp]
    \centering
    \subfloat[][\label{fig:wam_traj_costs}]{%
      \includegraphics[width=0.24\linewidth,trim={0.1cm 0cm 2.05cm 0.5cm},clip]{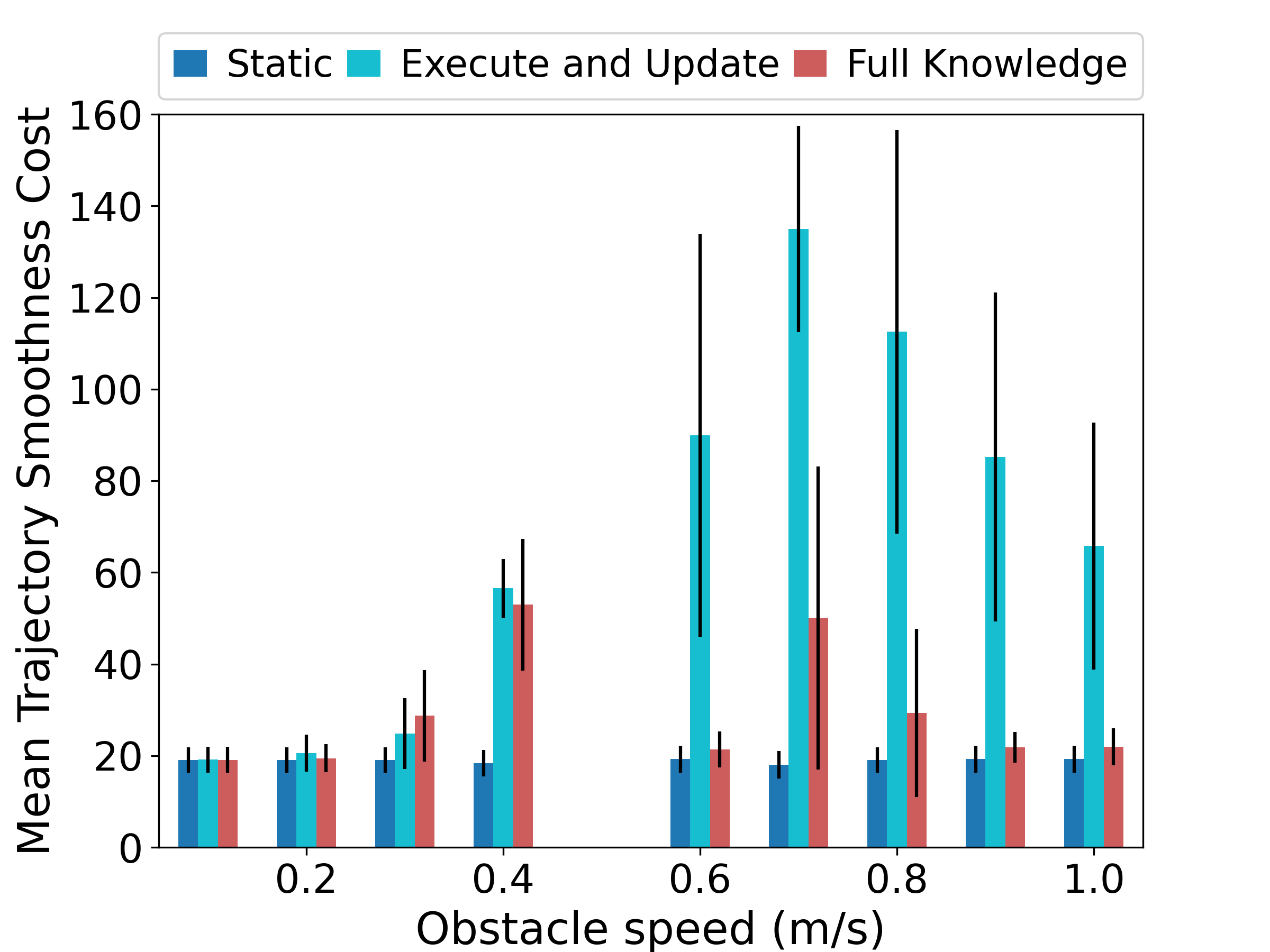}}
    \hfill
    \subfloat[][\label{fig:wam_traj_collisions}]{%
        \includegraphics[width=0.24\linewidth,trim={0.1cm 0cm 2.05cm 0.5cm},clip]{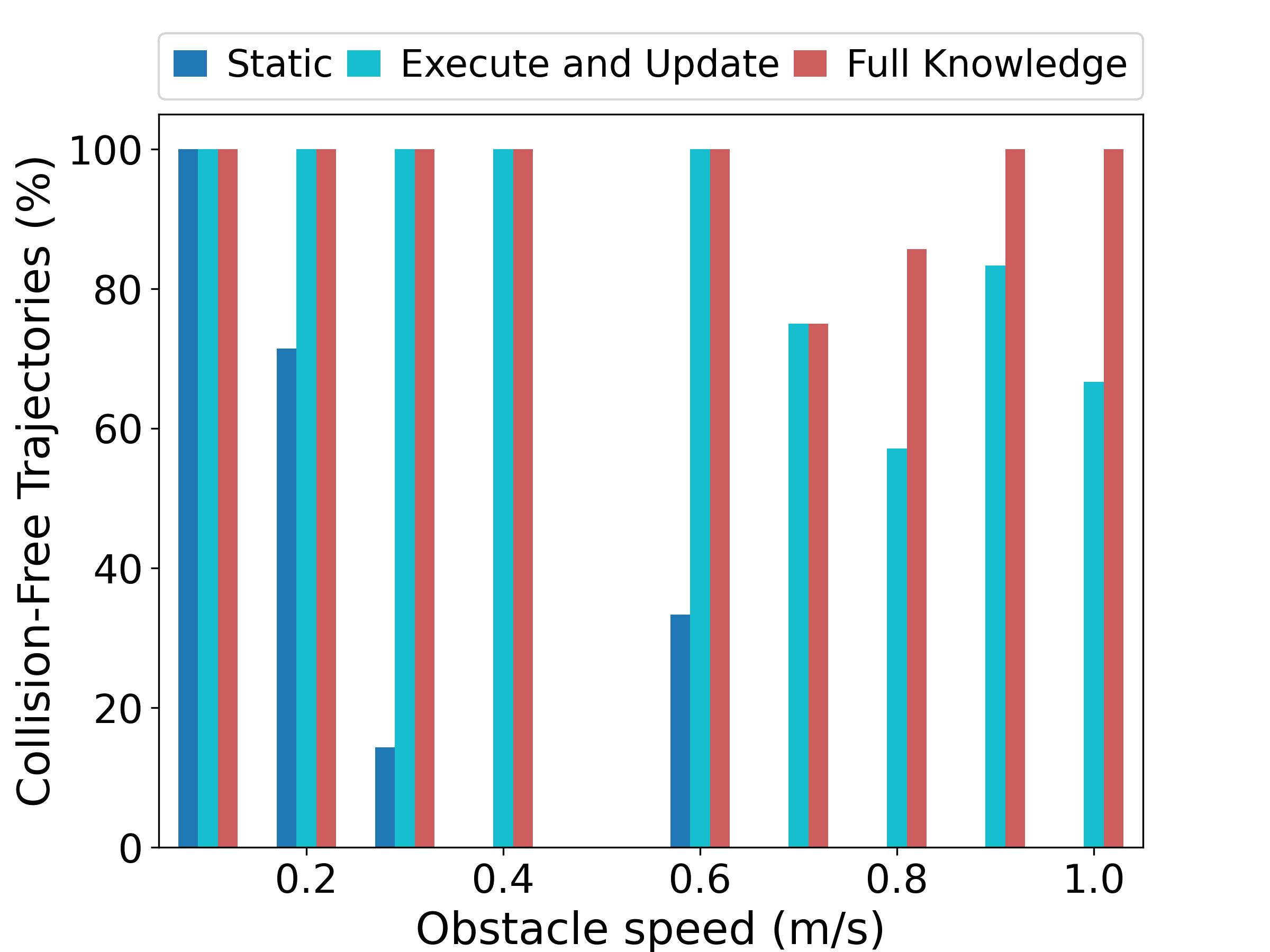}}
    \hfill
    \subfloat[][\label{fig:hsr_traj_costs}]{%
      \includegraphics[width=0.24\linewidth,trim={0.1cm 0cm 2.05cm 0.5cm},clip]{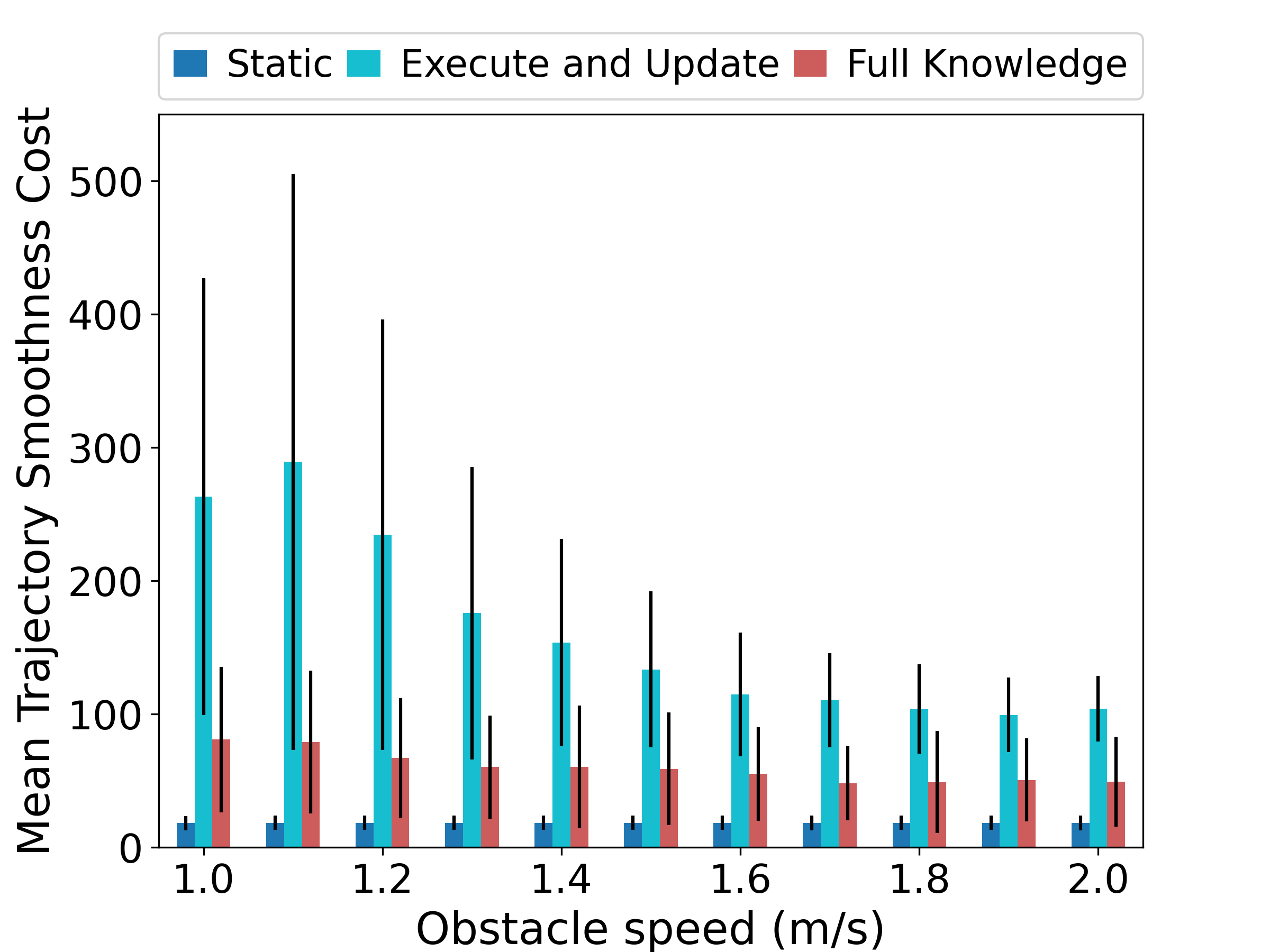}}
    \hfill
    \subfloat[][\label{fig:hsr_traj_collisions}]{%
        \includegraphics[width=0.24\linewidth,trim={0.1cm 0cm 2.05cm 0.5cm},clip]{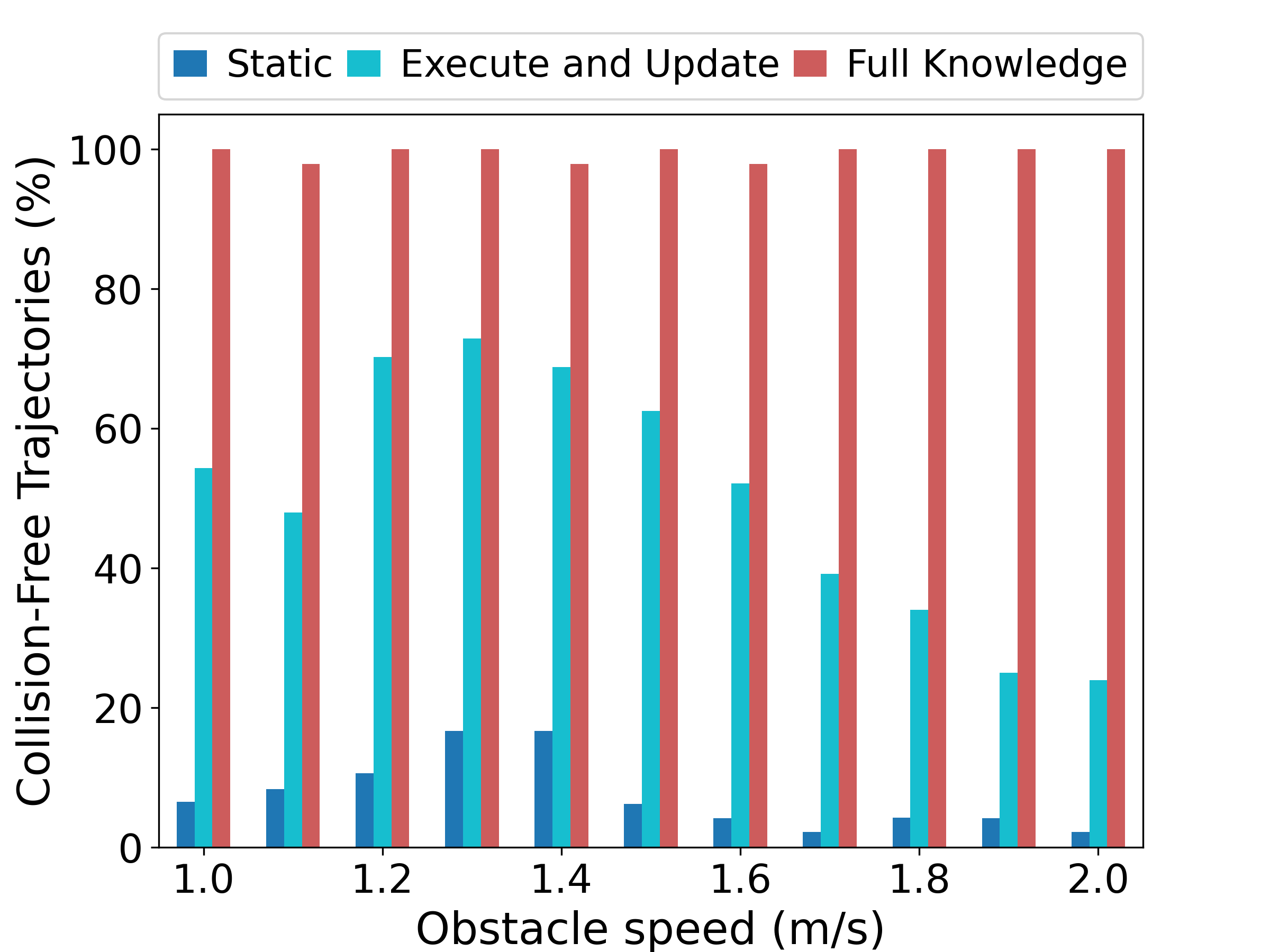}}
    \hfill
  \caption{Results for experiments conducted on the WAM arm in the `single moving pillar' setup are shown in Figures \ref{fig:wam_traj_costs} and \ref{fig:wam_traj_collisions}. Results for the Panda arm avoiding a moving HSR are shown in \ref{fig:hsr_traj_costs} and \ref{fig:hsr_traj_collisions}. We see that in all cases, using full knowledge of obstacle trajectories provides at least an equal chance of finding a collision-free trajectory using GPMP2; for most obstacle speeds, the improvement is significant.}
  \label{fig:dynamic_results} 
\end{figure*}

The performance of the three scenarios described in Section \ref{sect:dynamic_gpmp2}, on the `single moving pillar' dataset, are presented in Figure~\ref{fig:dynamic_results}. The figure shows an absence of data for the obstacle speed of \SI{0.5}{\metre\per\second}; at \SI{0.5}{\metre\per\second}, the obstacle intersects the goal configurations at the end of the \SI{3}{\second} time horizon, resulting in inevitable collisions. Therefore, we exclude results for \SI{0.5}{\metre\per\second} from our analysis. Of 69 planning results, we exclude three for which none of the planning scenarios found a collision-free trajectory.

Our results strongly support the incorporation of future obstacle motions into motion planning. Figure~\ref{fig:dynamic_results} shows that in most cases, including the obstacle trajectory motion into the planning significantly increases the likelihood of both finding a collision-free trajectory and reducing the jerk associated with the trajectory. Furthermore, the results motivate the use of our proposed framework in which we generate predicted ESDFs in real-time.

We note here that our implementation of a dynamic obstacle factor in GPMP2 incurs a cost of \SI{\approx2}{\milli\second} to be replaced by an obstacle factor with a different SDF.

\subsection{Closed-Loop Implementation}
\begin{figure*}
    \centering
    \subfloat[][Our method - including prediction\label{fig:prediction_joint_plot}]{%
        \includegraphics[width=0.5\columnwidth]{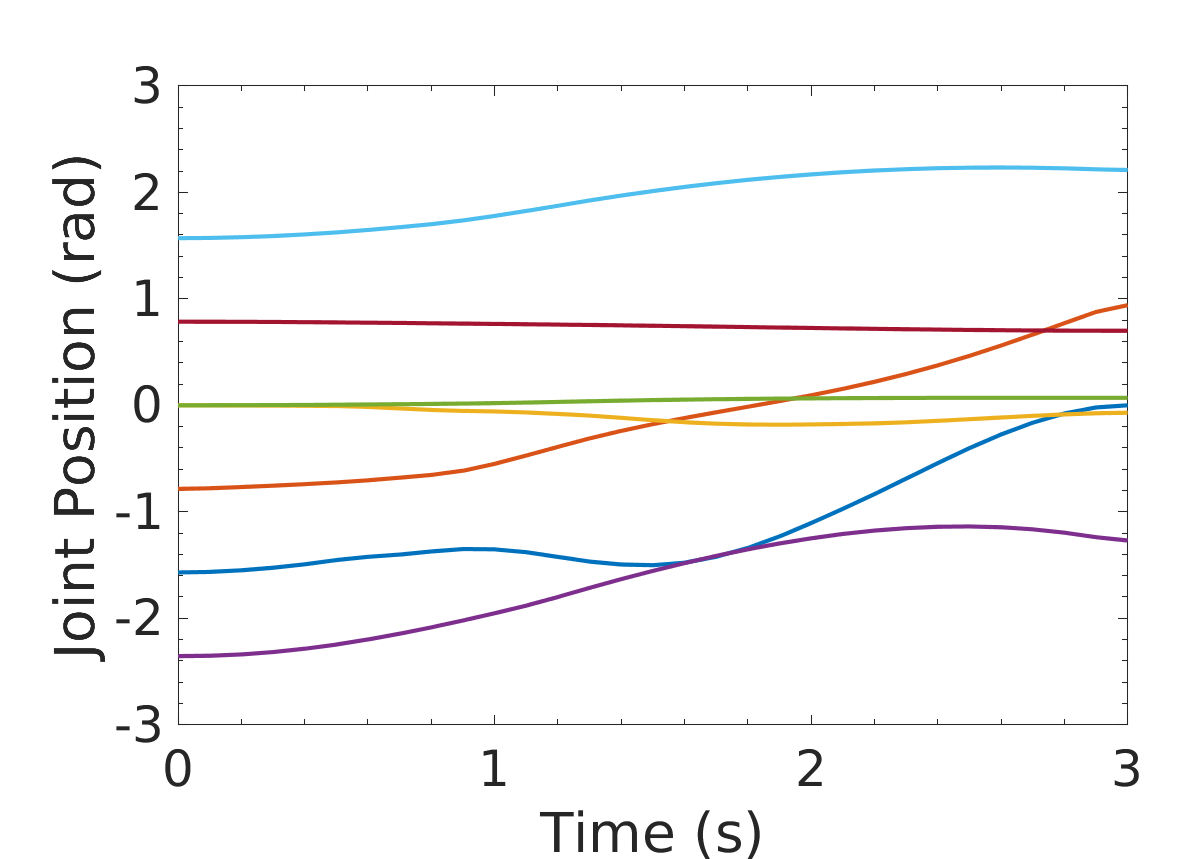}   \includegraphics[width=0.5\columnwidth]{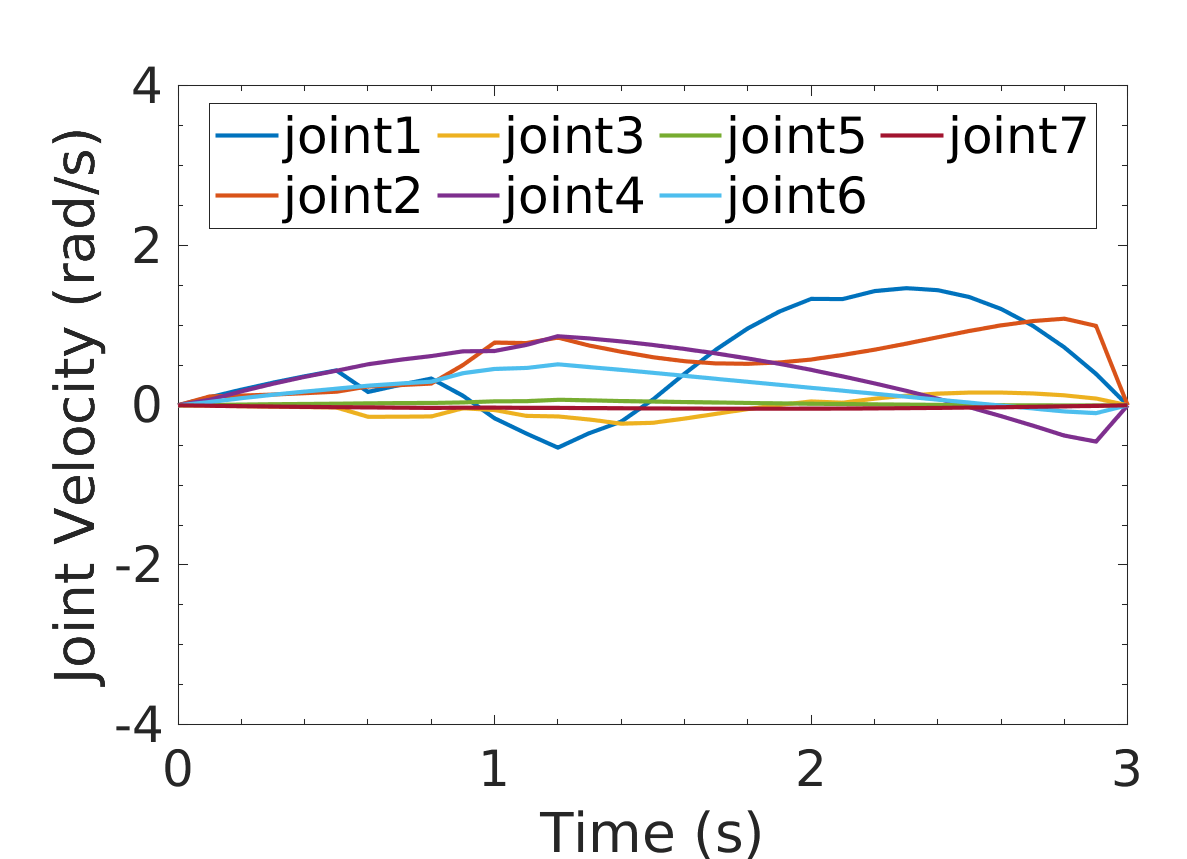}
        }
    \subfloat[][Execute and update\label{fig:exec_update_joint_plot}]{%
        \includegraphics[width=0.5\columnwidth]{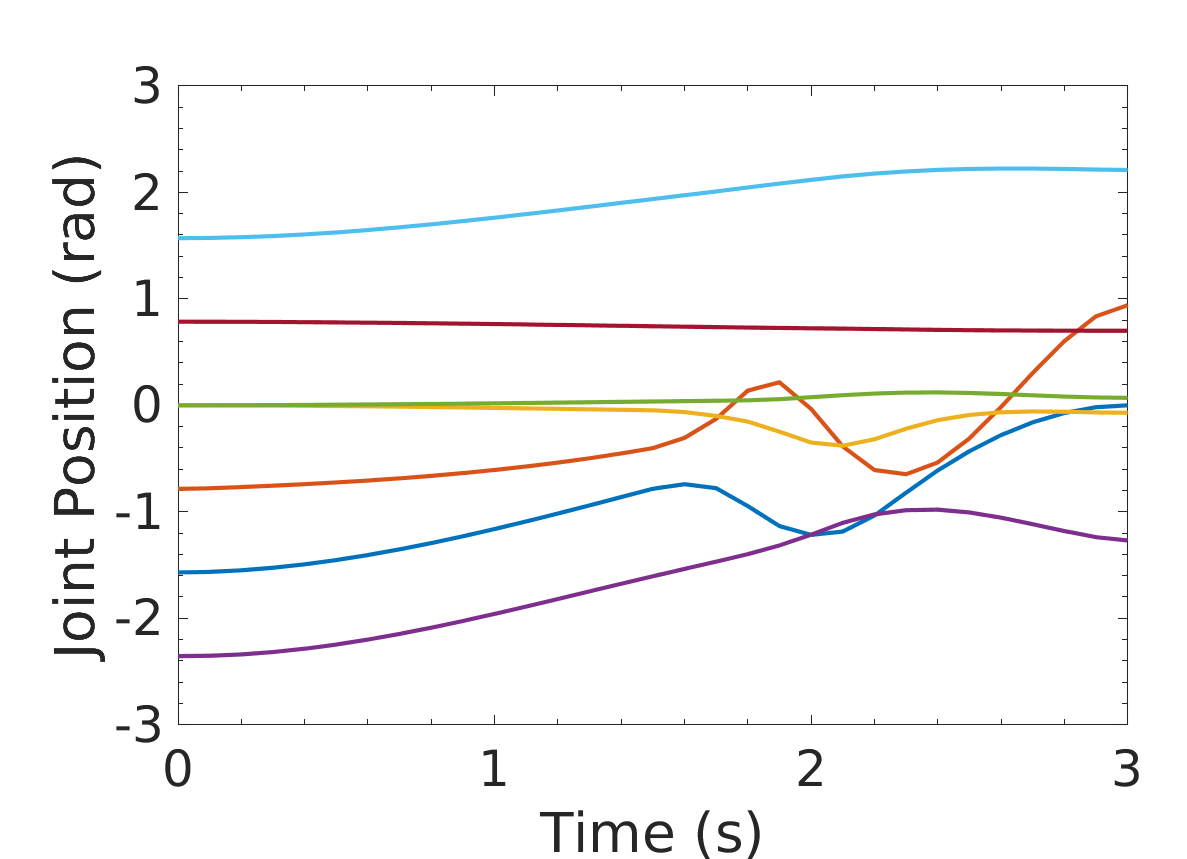} \includegraphics[width=0.5\columnwidth]{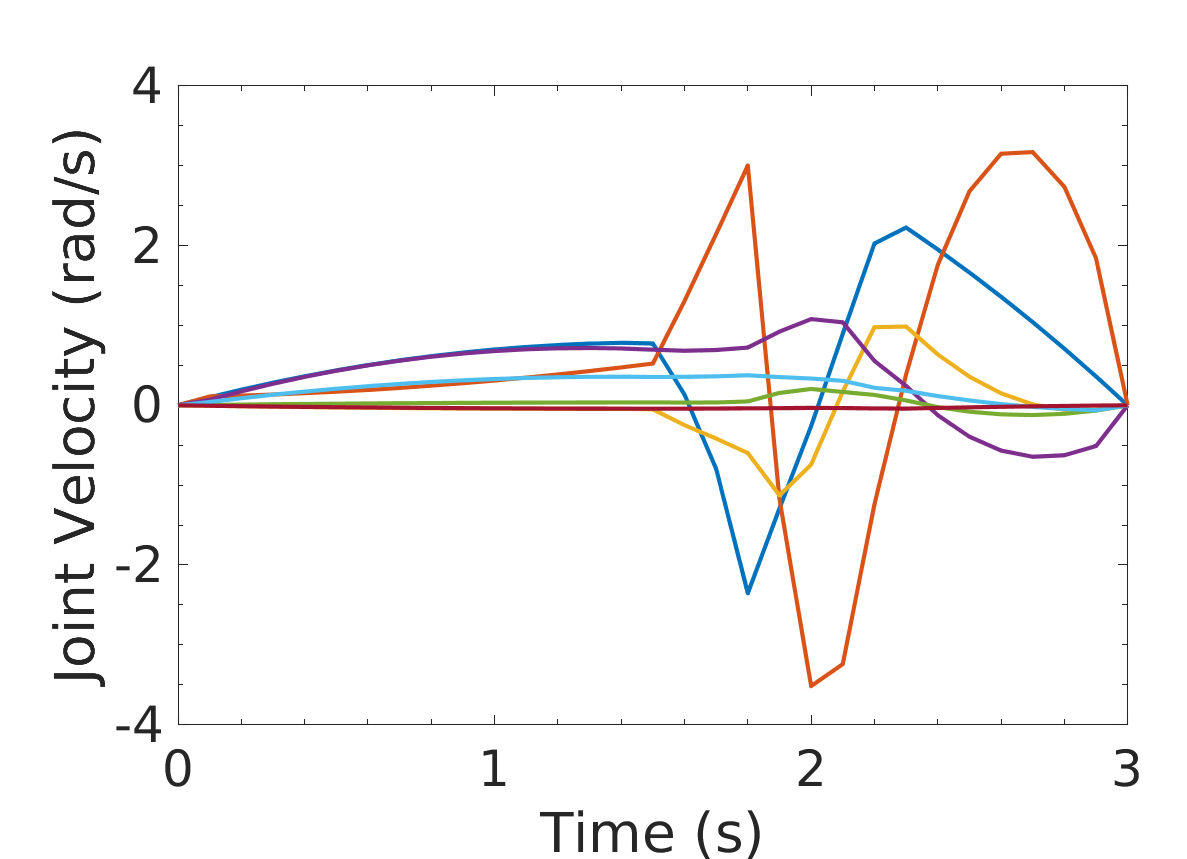}
        }
    \caption{Example trajectories for joint position and velocity. Our results highlight that the inclusion of environment prediction in motion planning can lead to smoother trajectories which exhibit less erratic movements.}
    \label{fig:joint_position_plot}
\end{figure*}

Our closed-loop implementation was successful in providing re-optimised trajectories fast enough for a Panda arm to avoid a moving HSR robot while reaching across a gap between two tables. Figure~\ref{fig:intro} illustrates the trajectory taken in each of the three cases. The `static' case would result in a collision and the `execute and update' would exhibit erratic movements. Figure~\ref{fig:joint_position_plot} shows a comparison of the resultant joint position trajectories for the `execute and update' against our composite prediction method. The results lend further support that by including \textit{predicted SDFs} into motion planning, we can obtain smoother robot trajectories. 

 This experiment was conducted using a $96 \times 96 \times 96$ workspace with \SI{4}{\centi\metre} resolution. The parameters for the motion planner were $n_{int} = 1$, $\sigma_{obs} = 0.05$, $\epsilon = \SI{0.3}{\metre}$. A time horizon of \SI{3}{\second} was given, with variable factors time-indexed in steps of $\delta_t = 0.1\si{\second}$, resulting in 31 support states. The first iteration of the control loop results in the slowest update frequency because more future factors need to be updated and more predictions are generated. Averaged over 100 runs, our implementation runs at \SI{3.4}{\hertz} in the early iterations, with a standard deviation of \SI{0.1}{\hertz}. In the later iterations, we achieve an update frequency of \SI{\approx10}{\hertz}. We demonstrate the resultant trajectories for each of the three cases in the accompanying video. \footnote{\url{https://youtu.be/vJH8qBRMedw}}


\subsection{Hardware Experiments} 
\begin{figure*}[htbp]
    \centering
    
    \subfloat[][\label{fig:start_predict}]{%
       \includegraphics[width=0.33\linewidth,trim={0cm 0.6cm 0cm 0cm},clip]{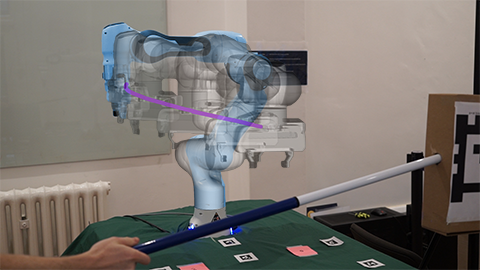}}
    \subfloat[][\label{fig:mid_predict}]{%
        \includegraphics[width=0.33\linewidth,trim={0cm 0.6cm 0cm 0cm},clip]{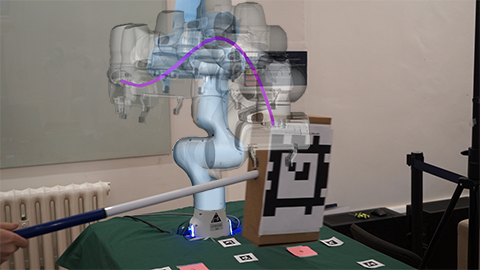}}
    \subfloat[][\label{fig:last_predict}]{%
        \includegraphics[width=0.33\linewidth,trim={0cm 0.6cm 0cm 0cm},clip]{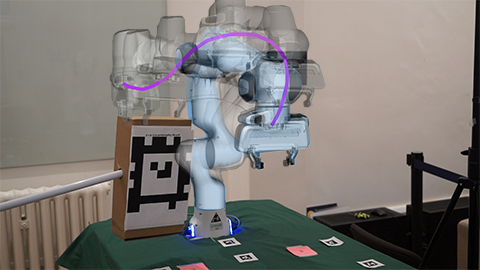}}

    \vfill

    \subfloat[][\label{fig:start_bench}]{%
       \includegraphics[width=0.33\linewidth,trim={0cm 0.6cm 0cm 0cm},clip]{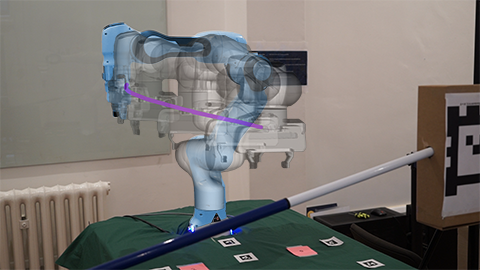}}
    \subfloat[][\label{fig:mid_bench}]{%
        \includegraphics[width=0.33\linewidth,trim={0cm 0.6cm 0cm 0cm},clip]{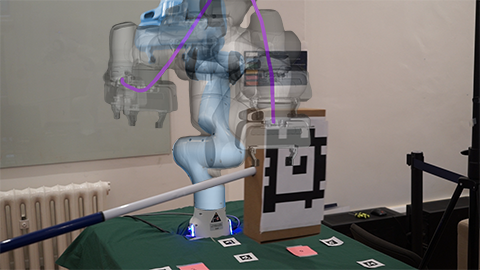}}
    \subfloat[][\label{fig:last_bench}]{%
        \includegraphics[width=0.33\linewidth,trim={0cm 0.6cm 0cm 0cm},clip]{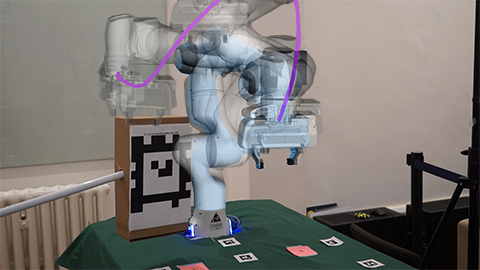}}
  \caption{\textit{Top row} - online motion planning using our approach of composite signed distance fields and prediction. \textit{Bottom row} -  using composite signed distance fields but without prediction. From left to right, the images show the planned trajectory for times $t = \SI{0}{\second}$, $t = \SI{2.5}{\second}$ and $t = \SI{5.0}{\second}$ (resultant trajectory). The blue highlight indicates the configuration of the robot at the current time-step while the purple line show the planned end-effector trajectory. The predictive component of our framework enables a smoother resultant trajectory, as shown in Figure~\ref{fig:last_predict}.}
  \label{fig:real_robot}
\end{figure*}
Our implementation performed well on the real robot arm, producing re-planned trajectories which successfully avoided the dynamic object in many cases. An example comparison against using composite SDFs but without prediction is shown in Figure~\ref{fig:real_robot}. By accounting for the predicted motion of the obstacle, the manipulator is able to avoid the obstacle by following a much smoother trajectory.

We found that while neglecting to predict the obstacle motion was also effective at avoiding obstacles due to the high re-plan frequency, the robot would exhibit erratic behaviours, in particular when the obstacle position coincided with the desired arm goal state.


 
 \vspace{-2.2pt}
\section{Discussion}
Our proposed framework gives compelling results in support of incorporating predicted obstacle trajectories into motion planning in time-configuration space. While we implemented a constant-velocity model to demonstrate our framework's novel ability to include obstacle trajectory predictions directly into SDF-based based motion planning, one can replace this with a more complex model that takes into account any prior information we have on the obstacle's trajectory, including its history. Exploring different prediction models for obstacle motion was beyond the scope of this work and as such we did not explore the avoidance of obstacles that follow complex motions, such as curved trajectories; we leave this for future work. Without using a more complex model, our current implementation is likely to fail in circumstances where the obstacle suddenly changes direction or stops.

A limitation of the proposed method is the potentially large memory requirements associated with assigning and storing different signed-distance fields for different time-indexed obstacle factors. For example, if a $300 \times 300 \times 300$ discretisation of the workspace is required for an application, our previous problem of a \SI{3}{\second} time horizon, discretised into 31 time-steps, will require \SI{\approx6.24}{\giga\byte} RAM.\footnote{A \texttt{double} is stored using \SI{8} bytes. $300^{3} \times 8 = 2.16 \times 10^8 \text{ bytes }  = \SI{206}{\mega\byte}$. $31 \times 206 \SI{ \approx 6.24}{\giga\byte}$} We thus encounter a trade-off; minimising the workspace size results in quicker computation and lower memory requirements, however, to monitor a larger workspace requires the resolution to be decreased. We believe that this method could be adapted in future work to perform the superposition on a query basis, rather than computing the composite signed-distance field for the entire workspace.

Generating composite predictions for each time-step, as well as the updating of the obstacle factors, are operations which seem suitable for parallelisation. By updating factors in parallel, we believe that significantly higher update rates could be achieved allowing for more interactive and responsive behaviour.

We have demonstrated benefits of predicted composite SDFs that are independent of state-of-the-art SDF computing techniques. In our future work, we will evaluate different methods that can leverage our technique and framework, particularly parallel implementations. We also note that while pre-computing a static scene is acceptable for a fixed-base robot arm, it is less useful in the context of a mobile robot performing tasks in an unmapped region of space. To address this, we are working on integrating our approach with ESDFs that are updated from live point-cloud data.


 
 \vspace{-2.2pt}
\section{Conclusion}
This paper explored the application of composite signed-distance fields to motion planning in dynamic environments. We first exploited the speed of the \texttt{min} operation for composition of SDFs, and secondly, that signed-distance field representations must only be accurate up to a specified $\epsilon$ for motion planning problems. We show that composite SDFs can be used to provide significant speed-up for generating SDFs when accounting for moving obstacles.
We investigated motion planning with a GPMP2 implementation which uses dynamic obstacles factors, enabling the planner to account for moving obstacles. Over a range of tasks and two different robot platforms, our results show that by incorporating predicted obstacle trajectories, we can significantly reduce the smoothness cost of trajectories and the rate of collisions. 
We leveraged composite SDFs and a dynamic GPMP2 implementation to present a novel framework which exploits the sparsity of the workspace and the compositional nature of signed-distance fields to generate real-time predictions of the workspace SDF, \textit{predicted signed-distance fields}. We verified our approach on a real 7-DoF Panda arm with obstacle tracking performed using fiducial sensing, demonstrating that it can plan trajectories in dynamic environments and successfully avoid moving obstacles.

 \vspace{-2.2pt}
\section{Acknowledgments}
This research was supported by (1) the UK Engineering and Physical Sciences Research Council (EPSRC) through the University of Oxford Centre for Doctoral Training, Autonomous Intelligent Machines and Systems (AIMS, grant references EP/L015897/1 and EP/R512333/1), the UK RAI Hub for Offshore Robotics for Certification of Assets (ORCA, grant reference EP/R026173/1), and the EPSRC grant Robust Legged Locomotion (EP/S002383/1), as well as (2) the European Commission under the Horizon 2020 project Memory of Motion (MEMMO, project ID: 780684).

\pagebreak
\small
\bibliography{main}

\end{document}